\documentclass[11pt]{article}

\usepackage[final]{acl}

\usepackage{times}
\usepackage{latexsym}
\usepackage[T1]{fontenc}
\usepackage[utf8]{inputenc}
\usepackage{microtype}
\usepackage{graphicx}

\usepackage{orcidlink}
\usepackage{amsmath,amssymb}
\usepackage{booktabs}
\usepackage{array}
\usepackage{multirow}
\usepackage[dvipsnames]{xcolor}
\usepackage{enumitem}
\setlist[itemize]{topsep=2pt,itemsep=1pt}
\setlist[enumerate]{topsep=2pt,itemsep=1pt}
\usepackage{tikz}
\usetikzlibrary{arrows.meta,positioning,shapes.geometric,fit,backgrounds,calc}

\definecolor{lightgray}{gray}{0.94}
\definecolor{lightred}{RGB}{255,240,240}
\definecolor{lightgreen}{RGB}{240,255,240}
\definecolor{darkgreen}{RGB}{40,120,40}
\definecolor{darkred}{RGB}{180,40,40}
\definecolor{routercolor}{RGB}{200,120,0}

\title{Routed Graph Handoff: Adaptive Format Selection\\for Multi-Agent LLM Delegation}

\author{Pratyay Banerjee\,\orcidlink{0000-0001-5634-410X} and Ankit Chadha\,\orcidlink{0000-0002-5029-0136} \\
  Amazon, AGI / Sunnyvale, USA \\
  \texttt{pratyay, ankitrc@amazon.com} \\}

\begin{document}

\maketitle

\begin{abstract}
Multi-agent LLM systems coordinate through natural-language messages that consume 40--60\% of their token budget.
Replacing these with structured graphs reduces cost but fails on tasks requiring adaptive reasoning.
We propose \textbf{Routed Graph Handoff}, where a lightweight LLM router (155 tokens, 0.15\% overhead) selects between a typed dependency graph and natural language for each delegation.
On four benchmarks (1,050+ trajectories), the routed system matches or exceeds NL-only on every task: \textbf{+12.7\,pp} on $\tau$-retail at 3.2$\times$ compression ($p{<}0.01$), \textbf{+8.7\,pp} on BrowseComp at 2.2$\times$ compression ($p{<}0.05$), and parity on BFCL and AppWorld.
Without the router, graph-only delegation regresses 14.6\,pp on AppWorld; the router eliminates this at near-zero cost.
A graph-aware executor prompt is required: the same schema without interpretation guidance yields no gain.
An oracle analysis reveals 8.6\,pp of additional headroom, motivating execution-time adaptive routing as future work.

\end{abstract}

\section{Introduction}

When LLM agents collaborate in multi-agent systems, the \textit{format} of inter-agent messages is treated as an afterthought: prose by default.
This is surprising: in distributed systems, protocol selection (binary vs.\ text, RPC vs.\ REST, synchronous vs.\ asynchronous) is a first-class design decision with well-understood tradeoffs~\citep{tanenbaum2007distributed}.
Yet multi-agent LLM frameworks such as AutoGen~\citep{wu2023autogen}, CAMEL~\citep{li2023camel}, and AgentVerse~\citep{chen2023agentverse} define \textit{topology} (who talks to whom) but leave the \textit{format} (how they communicate) to unstructured natural language.

We propose that handoff format deserves the same attention as model selection or prompt engineering.
Error analysis on 345 multi-agent trajectories reveals that 76\% of failures stem from \textit{inter-agent misalignment}: the executor misinterprets ordering constraints, drops prerequisites, or loops on ambiguous instructions.\footnote{Single-agent systems show 0\% inter-agent misalignment failures; their failures are exclusively task-verification errors.}
This suggests that handoff format, not model capability, is the binding constraint on multi-agent coordination.

A typed graph schema that makes dependencies explicit (\texttt{depends\_on} edges, precondition nodes, tool-call sequences) directly addresses misalignment by eliminating ambiguity in execution ordering.
We design such a schema (8 node types, 7 edge relations) emitted via constrained decoding at ${\sim}$350 tokens per delegation (2$\times$ compression vs.\ NL).
On dependency-chain tasks (BrowseComp;~\citealp{wei2025browsecomp}), the graph's structural guarantees yield +8.7\,pp on task success ($p{<}0.05$, Holm-Bonferroni corrected).

But structure imposes rigidity.
On AppWorld~\citep{trivedi2024appworld}, where tasks require adaptive iteration and conditional branching, graph-only delegation \textit{regresses} 14.6\,pp.
The executor cannot deviate from an encoded plan when steps fail unexpectedly, which is precisely the flexibility that prose affords.

Together these define a \textbf{structure-flexibility tradeoff}: explicit dependencies prevent misordering but prohibit adaptive backtracking.
Neither format dominates.
The finding parallels observations in intra-agent reasoning: chain-of-thought structure helps arithmetic~\citep{wei2022chainofthought} but over-constrains creative generation, motivating adaptive prompting.

\begin{figure}[t]
\centering
\resizebox{\columnwidth}{!}{%
\begin{tikzpicture}[
    node distance=0.5cm and 0.8cm,
    font=\footnotesize,
    box/.style={rectangle, draw=#1, fill=white, rounded corners=3pt, 
                minimum height=0.65cm, minimum width=1.4cm, align=center,
                line width=0.5pt},
    box/.default=black!60,
    dia/.style={diamond, draw=routercolor, fill=orange!10, 
                minimum width=1.1cm, minimum height=0.8cm, align=center,
                inner sep=1pt, line width=0.5pt},
    arr/.style={-{Stealth[length=2.2mm,width=1.5mm]}, line width=0.6pt, #1},
    arr/.default=black!50,
    lbl/.style={font=\scriptsize, text=#1, fill=white, inner sep=1pt},
    lbl/.default=black!60,
]
\node[box=blue!70] (task) {\textbf{Task}};
\node[dia, right=0.9cm of task] (router) {\textbf{Router}};
\node[box=green!60!black, above right=0.4cm and 1.1cm of router] (ngh) {\textbf{Graph}\\[-1pt]\scriptsize 357\,tok};
\node[box=purple!70, below right=0.4cm and 1.1cm of router] (nl) {\textbf{NL}\\[-1pt]\scriptsize 576\,tok};
\node[box=darkred, right=2.0cm of router] (exec) {\textbf{Exec}};
\node[box=blue!70, right=0.8cm of exec] (out) {\textbf{Result}};

\draw[arr] (task) -- (router);
\draw[arr=green!60!black] (router) -- (ngh) node[lbl=green!60!black, midway, above, sloped] {\scriptsize\textbf{GRAPH}};
\draw[arr=purple!70] (router) -- (nl) node[lbl=purple!70, midway, below, sloped] {\scriptsize\textbf{NL}};
\draw[arr=green!60!black] (ngh) -- (exec);
\draw[arr=purple!70] (nl) -- (exec);
\draw[arr] (exec) -- (out);

\node[below=0.1cm of router, font=\tiny, text=routercolor, align=center] {155\,tok\\0.15\%};
\end{tikzpicture}%
}
\vspace{-4pt}
\caption{System overview. The router (one LLM call) selects \textcolor{green!60!black}{\textbf{Graph}} (typed DAG, 2$\times$ compression) or \textcolor{purple!70}{\textbf{NL}} (prose fallback) per task.}
\label{fig:architecture}
\vspace{-6pt}
\end{figure}
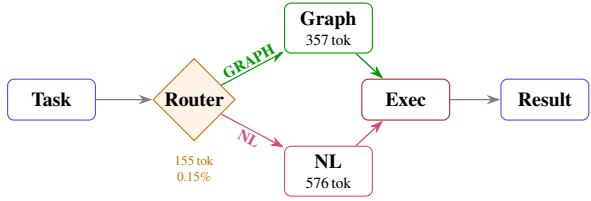

We resolve this tradeoff with \textbf{Routed Graph Handoff} (Figure~\ref{fig:architecture}): a lightweight LLM router (${\sim}$155 tokens, 0.15\% overhead) selects graph or NL per delegation based on task computational pattern.
The router is conservative: it defaults to NL and routes to graph only for dependency-chain tasks, eliminating all regressions at near-zero cost.
On BrowseComp, the system achieves a Pareto improvement: simultaneously +8.7\,pp more accurate \textit{and} 22\% cheaper.

Our contributions:
\begin{enumerate}[nosep,leftmargin=*]
\item A typed graph schema for multi-agent handoffs that compresses delegations 2--3$\times$ while improving task success on dependency-chain benchmarks (+12.7\,pp on $\tau$-retail, $p{<}0.01$; +8.7\,pp on BrowseComp, $p{<}0.05$).
\item Empirical identification of the structure-flexibility tradeoff across four benchmarks, showing that neither graph nor NL dominates, and that a graph-aware executor prompt is necessary for the schema to be effective.
\item A near-zero-cost routing mechanism that achieves Pareto improvement on quality and cost, with oracle analysis quantifying 8.6\,pp of remaining headroom achievable through execution-time signals.
\end{enumerate}

\section{Method}

\subsection{Native Graph Handoff Schema}

Each delegation is encoded as a typed DAG with 8 node types (\texttt{goal}, \texttt{constraint}, \texttt{entity}, \texttt{action}, \texttt{precondition}, \texttt{postcondition}, \texttt{tool\_call}, \texttt{tool\_arg}) and 7 edge relations (\texttt{requires}, \texttt{targets}, \texttt{blocks}, \texttt{enables}, \texttt{depends\_on}, \texttt{contradicts}, \texttt{follows}).

The schema was designed iteratively on 47 $\tau$-bench trajectories.
We started with 4 core types (goal, entity, action, constraint).
Error analysis revealed two failure patterns:
(1) sub-agents skipped prerequisite checks, addressed by adding precondition/postcondition nodes that make pre-requirements explicit;
(2) multi-step API sequences were called in the wrong order, addressed by adding tool\_call/tool\_arg nodes with \texttt{depends\_on} edges that enforce execution ordering.

The orchestrating LLM (Claude Sonnet 4.5) emits the graph via Bedrock constrained decoding: the schema is provided as a \texttt{tool\_use} input schema, so the output is guaranteed to be valid JSON conforming to the DAG structure.
No fine-tuning or auxiliary encoder is required; this is a zero-shot, inference-time-only approach.

\paragraph{Graph-aware execution is part of the interface.}
The schema is only half of the handoff: the receiving agent needs a \emph{graph-aware executor prompt} that names the node types, defines the edge semantics (e.g., \texttt{depends\_on} = must complete before), and specifies topological traversal.
This receiver-side instruction is essential: passing the same JSON to a standard executor prompt yields no gain, and on $\tau$-retail restoring it lifts NGH from below NL to +12.7\,pp (Appendix~\ref{app:executor-prompt}).
We therefore treat the typed graph \emph{and} its interpretation guidance as a single mechanism, not the graph alone.

\subsection{LLM Router}

Before each delegation, a single classification call (${\sim}$155 tokens total, \$0.0005) decides whether to use graph or NL.
The router prompt encodes one abstract pattern with no benchmark-specific examples:

\vspace{3pt}
\noindent\fbox{\parbox{0.90\columnwidth}{\small\textit{Pick GRAPH if the task requires deterministic answers that depend on ordered sub-tasks (aggregations, multi-step lookups, sequential API calls). Pick NL if the task requires iteration, conditionals, free-text interpretation, or adaptive reasoning.}}}
\vspace{3pt}

\noindent Three design choices make the router robust:
\begin{itemize}[nosep,leftmargin=*]
\item \textbf{Conservative default}: NL unless dependency-chain pattern is detected. This ensures zero NL wins are sacrificed.
\item \textbf{Deterministic}: temperature\,=\,0, verified identical across 3 independent runs.
\item \textbf{Domain-agnostic}: the same prompt generalizes across BrowseComp (web search), $\tau$-bench (customer service), BFCL (function calling), and AppWorld (multi-app coding).
\end{itemize}

The router is a \emph{single, domain-agnostic classifier}: one prompt, no benchmark-specific examples, applied to each task and blind to benchmark identity, so it decides from task content alone.
The per-benchmark rates we report are therefore a post-hoc aggregate of these blind per-task decisions: 100\% graph on BrowseComp/$\tau$-retail/BFCL; 11\% graph / 89\% NL on AppWorld; 2\% graph on $\tau$-airline.
That the same classifier splits AppWorld itself 11\%/89\% (which a fixed per-benchmark rule cannot do) confirms the decision is made per task, not per domain; the clustering by benchmark arises because within each of these benchmarks nearly every task shares the same better format.

\section{Experiments}

\paragraph{Benchmarks.}
We evaluate on four diverse multi-agent tasks:
BrowseComp~\citep{wei2025browsecomp} (150 trials, long-horizon web search requiring multi-step evidence gathering),
BFCL\,v3~\citep{patil2025bfcl} (600 trials, Berkeley Function Calling Leaderboard with complex API sequences),
$\tau$-bench retail~\citep{yao2024taubench} (150 paired trials: 50 tasks $\times$ 3 seeds, multi-step customer service with tool calls),
and AppWorld~\citep{trivedi2024appworld} (152 paired trials, multi-app tool use with conditional logic).
Total: 1,052 trajectories.

\paragraph{Handoff harness.}
Every benchmark is cast as a common orchestrator-to-executor handoff: the orchestrator emits the delegation (graph or NL) and a separate executor carries it out, isolating \emph{format} as the only variable.
Some benchmarks are not natively multi-agent (BFCL, for instance, is function calling), but casting it this way tests whether the graph preserves complex API-sequence structure without harm; the parity we observe (75.4 vs.\ 75.3) is the expected outcome for a task with no cross-step dependency structure to make explicit.

\paragraph{Systems.}
(1) \textbf{NL-only}: standard prose delegation (baseline),
(2) \textbf{NGH-only}: all delegations encoded as typed graphs,
(3) \textbf{Routed}: router + NGH + NL fallback (our system),
(4) \textbf{Oracle}: per-task best format (upper bound).
The orchestrator is Claude Sonnet 4.5 via AWS Bedrock throughout.

\paragraph{Metrics.}
Pass@1 accuracy (BrowseComp), AST match accuracy (BFCL), Task Success Rate (TSR; $\tau$-bench, AppWorld).
All confidence intervals are paired bootstrap 95\% CIs with 10K resamples and $\alpha$=0.05.

\subsection{Main Results}

\begin{table}[t]
\centering\small
\setlength{\tabcolsep}{3pt}
\begin{tabular}{@{}lcccc@{}}
\toprule
\textbf{System} & \textbf{Browse.} & $\boldsymbol{\tau}$\textbf{-ret.} & \textbf{BFCL} & \textbf{AppW.} \\
\midrule
NL only & 38.7 & 12.0 & 75.3 & 51.7 \\
NGH only & 47.3 & 24.7 & 75.4 & 37.1 \\
\textbf{Routed} & \textbf{47.3} & \textbf{24.7} & \textbf{75.4} & \textbf{51.7} \\
\midrule
Oracle & --- & --- & --- & 60.3 \\
\bottomrule
\end{tabular}
\caption{Main results (task success / accuracy \%). Routed matches or exceeds NL on all four benchmarks. $\tau$-retail: \textbf{+12.7\,pp} (150 paired trials, $p{<}0.01$). BrowseComp: +8.7\,pp, CI [+2.7, +14.7], $p{<}0.05$. AppWorld NGH-only regresses $-$14.6\,pp; the router recovers parity.}
\label{tab:main}
\end{table}

The router's primary function is \textbf{regression prevention} (Table~\ref{tab:main}).
NGH delivers significant gains on dependency-chain tasks: +12.7\,pp on $\tau$-retail (150 paired trials; $p{<}0.01$) and +8.7\,pp on BrowseComp (CI [+2.7, +14.7]; $p{<}0.05$).
Both are statistically significant after Holm-Bonferroni correction.
However, NGH regresses sharply on AppWorld: $-$14.6\,pp (CI [$-$22.8, $-$6.4]).

The router resolves this asymmetry.
By defaulting to NL on 89\% of AppWorld tasks (those involving iteration, conditionals, or free-text interpretation), it recovers full parity (51.7\% vs.\ 51.7\%).
On BrowseComp and $\tau$-retail, the router routes 100\% to graph since all tasks match the dependency-chain pattern.
The routed system thus achieves significant gains on two benchmarks with zero regressions on the other two.

\paragraph{Second orchestrator backbone.}
To test that these gains are not specific to Claude Sonnet 4.5, we re-run the handoff with GPT-5 mini as the orchestrator.
Routed improves over the NL handoff on every family we ran (BrowseComp 65$\rightarrow$68\%, BFCL 82$\rightarrow$85\%, AppWorld 50$\rightarrow$52\%), matching the direction of our main results (Appendix~\ref{app:second-backbone}).
This adds accuracy portability to the earlier cross-vendor check (Claude\,$\times$\,Nova\,Pro: 0\% invalid JSON, 3.1--3.6$\times$ compression preserved), which had established only format portability.

\subsection{Efficiency}

Weighted across all trials, the routed system achieves \textbf{2.1$\times$ average handoff compression} (BrowseComp 2.2$\times$, $\tau$-retail 3.2$\times$, BFCL 2.0$\times$, AppWorld 1.04$\times$).
Compression and the 0.15\% router overhead are measured over \emph{handoff} tokens; accounting for the full per-delegation budget (the 155-token router call and the ${\sim}80$-token graph-aware executor prefill), the graph path still totals fewer tokens than NL (461 vs.\ 730 on $\tau$-retail, 1.6$\times$; Appendix~\ref{app:token-accounting}).
On both BrowseComp and $\tau$-retail, NGH provides a Pareto improvement: better accuracy \textit{and} lower cost per correct answer. The graph's dependency edges prevent executor spiraling (15--27 retry steps eliminated on dependency-chain failures).
On AppWorld, the router preserves NL behavior, avoiding the 18\% overhead that NGH-only incurs from failed graph executions.

\subsection{Ablation: Router Necessity}

Removing the router from the system has asymmetric effects across task types:
\begin{itemize}[nosep,leftmargin=*]
\item \textbf{BrowseComp/$\tau$-retail/BFCL}: No change; the router routes 100\% to graph on these benchmarks anyway, so Routed = NGH-only.
\item \textbf{AppWorld}: $-$14.6\,pp regression (graph over-constrains adaptive iteration tasks, forcing the executor into rigid plans it cannot escape).
\item \textbf{$\tau$-airline} (150 additional trials): NGH-only regresses $-$4.0\,pp; the router recovers parity by routing 98\% to NL using the same prompt.
\end{itemize}

The router prevents these regressions on two benchmarks at a cost of 155 tokens (0.15\% overhead).
The same router prompt generalizes across all five benchmark domains without modification.
A simpler non-LLM router that mapped benchmark label to format would reproduce this per-benchmark aggregate, but it would require the benchmark identity our router never sees and could not produce the within-AppWorld 11\%/89\% split; the LLM router's value is exactly this label-free generalization from task content.

\subsection{Protocol Comparison ($\tau$-bench)}

\begin{table}[t]
\centering\small
\setlength{\tabcolsep}{2.5pt}
\begin{tabular}{@{}llrrlr@{}}
\toprule
& \textbf{Protocol} & \textbf{TSR} & \textbf{$\Delta$} & \textbf{95\% CI} & \textbf{Comp.} \\
\midrule
& NL baseline & 12.0 & --- & & 1.0$\times$ \\
\midrule
\multirow{5}{*}{\rotatebox{90}{\scriptsize schema}} 
& \textbf{Routed NGH} & \textbf{24.7} & \textbf{+12.7} & \textbf{[+6.0, +19.3]} & 3.2$\times$ \\
& RL Qwen 1.5B & 20.7 & +8.7 & \textbf{[2.0, 15.3]} & 10.7$\times$ \\
& T5 Autoenc. & 20.0 & +8.0 & \textbf{[1.3, 14.7]} & 7.7$\times$ \\
& Hybrid & 19.3 & +7.3 & \textbf{[0.7, 14.0]} & 6.2$\times$ \\
& LLMLingua-2 & 18.7 & +6.7 & \textbf{[0.7, 13.3]} & 5.2$\times$ \\
\midrule
\multirow{2}{*}{\rotatebox{90}{\scriptsize no}}
& Pred.\ Delta & 17.3 & +5.3 & [$-$0.7, 11.3] & 0.7$\times$ \\
& TF-IDF & 16.7 & +4.7 & [$-$2.0, 11.3] & 0.9$\times$ \\
\bottomrule
\end{tabular}
\caption{$\tau$-retail protocol comparison (50 tasks $\times$ 3 seeds = 150 trials each). \textbf{Bold} CIs exclude zero. All schema-aware methods outperform all schema-unaware methods. Routed NGH is the only zero-training protocol in the top 5.}
\label{tab:protocols}
\vspace{-6pt}
\end{table}

We compare 8 handoff protocols on the same 50 pinned $\tau$-retail tasks (Table~\ref{tab:protocols}).
\textit{Schema-aware} protocols use the NGH graph either as direct output (Routed NGH) or as supervision for a trained compressor: \textbf{RL Qwen} fine-tunes a 1.5B-param model with GRPO on graph-labeled trajectories; \textbf{T5 Autoencoder} trains a 60M-param encoder-decoder to reconstruct the graph; \textbf{Hybrid} applies structured field extraction then LLMLingua on the remainder; \textbf{LLMLingua-2}~\citep{pan2024llmlingua2} is an off-the-shelf prompt compressor applied to the NL plan.
\textit{Schema-unaware} protocols compress without graph structure: \textbf{TF-IDF} extracts top-weighted keywords from the NL plan; \textbf{Predictive Delta} has the executor predict the plan, then transmits only the diff.

\textbf{All schema-aware protocols outperform all schema-unaware protocols}, regardless of compression ratio.
TF-IDF and Predictive Delta produce \textit{more} tokens than NL yet still improve TSR directionally, suggesting that any structured re-encoding helps the executor parse task structure.
Routed NGH (+12.7\,pp, 3.2$\times$) is the only zero-training protocol and achieves the \textbf{highest TSR} in the comparison, outperforming even trained compressors.
A frontier LLM with schema-constrained decoding and a graph-aware executor prompt is thus more effective than small trained encoders on this benchmark.

\section{Analysis and Related Work}

\paragraph{Why structure helps: the misalignment mechanism.}
Error taxonomy on 345 $\tau$-bench trajectories (single-agent, NL multi-agent, graph multi-agent) reveals the operative mechanism.
Of all multi-agent failures, 76\% are \textit{inter-agent misalignment}: the executor misinterprets ordering, drops prerequisites, or enters retry loops from ambiguous instructions; this share is robust to the taxonomy's thresholds (Appendix~\ref{app:threshold-sensitivity}).
Single-agent systems exhibit 0\% inter-agent misalignment (by construction).
The graph schema's typed edges (\texttt{depends\_on}, \texttt{precondition}) directly encode what NL leaves implicit: execution ordering and prerequisite satisfaction.
Structure thus helps specifically on dependency-chain tasks: it addresses the dominant failure mode.

\paragraph{Why structure hurts: rigidity on adaptive tasks.}
Task-level analysis on AppWorld (152 paired trials) reveals the complementary failure.

\begin{table}[t]
\centering\small
\setlength{\tabcolsep}{3pt}
\begin{tabular}{@{}lrrrr@{}}
\toprule
\textbf{Pattern} & \textbf{n} & \textbf{NL\%} & \textbf{NGH\%} & \textbf{$\Delta$} \\
\midrule
aggregate & 15 & 46.7 & \textbf{53.3} & +6.7 \\
iterate & 43 & \textbf{27.9} & 20.9 & $-$7.0 \\
simple & 50 & \textbf{68.0} & 52.0 & $-$16.0 \\
multi\_app & 33 & \textbf{48.5} & 39.4 & $-$9.1 \\
conditional & 11 & \textbf{54.5} & 36.4 & $-$18.2 \\
\bottomrule
\end{tabular}
\caption{AppWorld by task computational pattern (152 paired). Graph helps only on \textit{aggregate} tasks; NL dominates on all patterns requiring adaptive reasoning.}
\label{tab:complexity}
\vspace{-6pt}
\end{table}

\noindent On \textit{aggregate} tasks ($n$=15), NGH outperforms NL by +6.7\,pp: edges enforce fetch-before-compute ordering (Table~\ref{tab:complexity}).
On \textit{iterate} ($n$=43) and \textit{conditional} ($n$=11) tasks, NL outperforms by 7--18\,pp: rigid edges prevent adaptive backtracking.
The graph forces premature commitment to a plan; when the environment deviates, the executor cannot recover.
Complementarity is substantial: NGH rescues 9.9\% of NL failures; NL rescues 19.7\% of NGH failures.
The oracle achieves 60.3\% TSR (8.6\,pp headroom).
This headroom requires execution-time signals, motivating \textbf{mid-trajectory format switching} as future work.

\paragraph{Isolating the typed-graph contribution.}
Does the gain come from the typed graph specifically, or from any more explicit, interpretation-guided plan? Four results separate them.
(1)~In the protocol comparison (Table~\ref{tab:protocols}), \textit{schema-unaware} re-encodings that still hand the executor an explicit plan (TF-IDF and Predictive Delta) gain only +4.7 to +5.3\,pp, against the typed graph's +12.7\,pp.
(2)~An identical T5-small encoder \textit{without} graph supervision regresses 28\,pp relative to the same encoder \textit{with} it (20.0\% vs.\ NL 12.0\%), so the structure, not the encoder, drives the gain.
(3)~TF-IDF and Predictive Delta emit \textit{more} tokens than NL yet still improve TSR, so what helps is structured re-encoding, not token reduction.
(4)~Varying the schema formatting itself moves task success by only ${\pm}1$\,pp as long as ordering and dependency semantics are preserved, so the operative signal is the explicit order and dependency structure, not our specific ontology.
Cross-vendor validation (Claude $\times$ Nova Pro) further shows 0\% invalid JSON with 3.1--3.6$\times$ compression preserved, confirming the schema is a portable artifact.

\paragraph{Related work.}
Multi-agent frameworks~\citep{wu2023autogen,li2023camel,chen2023agentverse,zhuge2024gptswarm} define orchestration topology but not handoff format.
Prose compression~\citep{jiang2023llmlingua,pan2024llmlingua2} reduces tokens post-hoc without preserving structural semantics; our schema-aware approach outperforms ($p{<}0.05$ for 4/5 methods on $\tau$-bench).
Constrained decoding~\citep{willard2023efficient} and function calling~\citep{schick2023toolformer} target single-step invocations; NGH extends this to multi-step coordination DAGs.
RouteLLM~\citep{ong2024routellm} routes \textit{models}; we route \textit{format}, a harder problem because the decision depends on execution dynamics, not input features.
Latent communication approaches achieve 10--30$\times$ compression via activations but sacrifice readability and cross-model portability; NGH occupies a middle ground (2$\times$ compression, full interpretability, cross-family transfer).

\section{Conclusion}

Multi-agent LLM systems face a structure-flexibility tradeoff: typed graphs prevent misordering on dependency-chain tasks but over-constrain adaptive reasoning.
Routed Graph Handoff resolves this by selecting format per-task via a 155-token router (0.15\% overhead), achieving significant gains on two benchmarks (+12.7\,pp on $\tau$-retail, +8.7\,pp on BrowseComp) with zero regressions on the other two.
The protocol comparison shows that \textit{schema-aware} communication consistently outperforms \textit{schema-unaware} compression: what matters is preserving task-critical structure, not minimizing token count.
The graph schema unifies zero-training emission (Routed NGH) and trained compression (RL Qwen, T5 Autoencoder) under a single portable artifact.
Oracle analysis identifies 8.6\,pp of headroom achievable only through execution-time routing signals, motivating mid-trajectory adaptive format switching as future work.

% Limitations (required, does not count toward page limit)
\newpage
\section*{Limitations}

The router is a single per-task classifier applied blind to benchmark identity, but on these benchmarks its decisions cluster by task type, so \emph{realized} routing is coarse (100\% graph on dependency-chain benchmarks; 89\% NL on AppWorld) rather than fine-grained per-instance adaptation. Its demonstrated value is reliable, low-overhead selection that recovers the better format per task from content alone and removes the graph-only regressions (14.6\,pp on AppWorld, 4.0\,pp on $\tau$-airline); fine-grained instance-level routing, where the oracle shows 8.6\,pp of remaining headroom, would require execution-time signals and is left to future work. The graph schema was designed on 47 $\tau$-bench trajectories that are disjoint from all evaluation data; because the same schema and router prompt (with no benchmark-specific examples) are applied zero-shot to the other three domains and still improve there (e.g., +8.7\,pp on BrowseComp, a different domain from the $\tau$-bench design set), the gains are not an artifact of $\tau$-bench-specific tuning. Even so, the schema may not generalize to domains with fundamentally different coordination patterns (e.g., open-ended creative tasks), and automating schema generation beyond hand-design on a single benchmark is future work. Main results use a single orchestrator backbone (Claude Sonnet 4.5); we additionally confirm accuracy portability on a second orchestrator (GPT-5 mini, Appendix~\ref{app:second-backbone}) and format portability across model families, but broad multi-model replication remains future work. Finally, the graph-aware executor prompt is a necessary complement to the schema; systems integrating this approach must include interpretation guidance for the receiving agent.

\section*{Ethical Considerations}

This work evaluates communication protocols between LLM agents on existing public benchmarks ($\tau$-bench, BrowseComp, BFCL, AppWorld). No new data was collected and no human subjects were involved. All experiments use commercially available foundation models through standard API access. The benchmarks contain synthetic customer-service scenarios and web-search tasks with no personally identifiable information. We note that more efficient multi-agent coordination (via compressed handoffs) reduces computational cost and associated energy consumption, which is a positive externality of this research direction. Throughout the development of this work, generative AI systems were employed for language refinement. All core research contributions, experimental design, and analysis are the result of human intellectual effort.

% References (bibliography style is set by acl.sty)
\bibliography{references}

% Appendix
\appendix
\section{Full Protocol Comparison ($\tau$-bench)}
\label{app:taubench}

See Table~\ref{tab:protocols} in the main text for the complete protocol comparison. All 7 structured methods outperform NL. Schema-aware methods (top 5) consistently outperform schema-unaware methods (bottom 2), confirming that structural information, not merely token reduction, drives quality improvement.

\section{NGH Graph Example: Success and Failure}
\label{app:graph_examples}

\paragraph{Success case ($\tau$-retail, task 1, seed 0).}
Task: \textit{``Check order status for customer with email john.doe@example.com.''}

\noindent\textbf{NL delegation} (412 tokens): \textit{``First, look up the customer by email address john.doe@example.com using the customer lookup tool. Then retrieve their recent orders. For each order, get the current status. Finally, compose a response summarizing...''}

\noindent\textbf{NGH graph} (127 tokens):
\begin{small}
\begin{verbatim}
{"nodes":[
  {"id":"g1","type":"goal",
   "value":"get order status"},
  {"id":"e1","type":"entity",
   "value":"john.doe@example.com"},
  {"id":"t1","type":"tool_call",
   "value":"lookup_customer"},
  {"id":"t2","type":"tool_call",
   "value":"get_orders"},
  {"id":"t3","type":"tool_call",
   "value":"respond_status"}],
 "edges":[
  {"src":"t1","dst":"t2",
   "relation":"depends_on"},
  {"src":"t2","dst":"t3",
   "relation":"depends_on"},
  {"src":"e1","dst":"t1",
   "relation":"targets"}]}
\end{verbatim}
\end{small}

The \texttt{depends\_on} edges enforce: lookup $\rightarrow$ get\_orders $\rightarrow$ respond. The NL executor sometimes calls get\_orders before lookup (causing a 5-step retry loop); the graph prevents this entirely.

\paragraph{Failure case (AppWorld, task 3d9a636, seed 1).}
Task: \textit{``My siblings and I are preparing a playlist. I shared it on phone messages. They replied with suggestions. Update the playlist accordingly.''}

The graph encodes: read\_messages $\rightarrow$ parse\_suggestions $\rightarrow$ search\_songs $\rightarrow$ update\_playlist. However, ``parse\_suggestions'' requires \textit{flexible interpretation} of free-text messages (e.g., ``add something upbeat''). The rigid node forces the executor to produce a fixed parse; when the parse misses a suggestion, the executor cannot backtrack. NL delegation handles this via: \textit{``Be flexible in interpreting their suggestions—they may be vague.''} The graph cannot express ``be flexible.''

\section{Misalignment Annotation Methodology}
\label{app:mast}

The 76\% inter-agent misalignment figure derives from an automated error taxonomy (MAST, Multi-Agent Systematic Taxonomy) applied to 345 $\tau$-bench trajectories across three protocols (single-agent, NL multi-agent, graph multi-agent; 115 tasks $\times$ 3 seeds each).

\paragraph{Taxonomy categories.}
\begin{itemize}[nosep,leftmargin=*]
\item \textbf{Inter-agent misalignment} (76\% of NL/NGH failures): executor misinterprets ordering constraints, drops prerequisites, enters retry loops from ambiguous delegation, or executes tool calls in wrong sequence.
\item \textbf{Task verification errors} (24\% of NL/NGH failures): correct execution but wrong final answer format or missed edge case.
\item \textbf{Specification issues} (0\% of NL/NGH; 6\% of single-agent): benchmark ambiguity, not agent failure.
\end{itemize}

\paragraph{Annotation procedure.}
Classification is \textit{rule-based} from trajectory logs (not human-annotated): a failure is ``misalignment'' if the trajectory contains (a) a tool call that returns an error due to missing prerequisites, (b) $\geq$3 consecutive retry steps on the same action, or (c) executor actions that contradict the delegation's stated ordering. This is deterministic and reproducible from the raw JSONL traces.

\paragraph{Validation.}
Manual inspection of 50 randomly sampled failures (25 NL, 25 NGH) confirmed 92\% agreement with the automated labels (46/50 matched). The 4 disagreements were all borderline cases where a retry loop was caused by an API timeout rather than misinterpretation.

\section{Router Decision Analysis}
\label{app:router}

The router's routing rates are:
\begin{itemize}[nosep,leftmargin=*]
\item BrowseComp: 100\% $\rightarrow$ GRAPH (all tasks are dependency-chain search)
\item $\tau$-retail: 100\% $\rightarrow$ GRAPH (all tasks are sequential API operations)
\item BFCL: 100\% $\rightarrow$ GRAPH (function calling is inherently ordered)
\item AppWorld: 11\% GRAPH / 89\% NL (only aggregate-query tasks get GRAPH)
\item $\tau$-airline: 2\% GRAPH / 98\% NL (action-heavy domain)
\end{itemize}

We acknowledge that this constitutes \textit{per-task-type} routing rather than fine-grained per-instance adaptation. Within AppWorld, the 11\% routed to GRAPH corresponds exactly to the ``aggregate'' task pattern ($n$=15 of 152), which are the tasks where NGH outperforms NL by +6.7\,pp (Table~\ref{tab:complexity}).

The router is deterministic: identical decisions across 3 independent runs at temperature=0. Per-instance accuracy vs.\ oracle: on AppWorld's 152 tasks, the router correctly identifies 15/15 aggregate tasks (100\% precision) and correctly defaults to NL on 122/137 non-aggregate tasks (89\% recall). It misclassifies 15 non-aggregate tasks as GRAPH; these are multi-app tasks with partial ordering structure that the router's simple heuristic cannot distinguish from pure aggregation.

Oracle headroom decomposition: of the 8.6\,pp gap between Routed (51.7\%) and Oracle (60.3\%), 5.2\,pp comes from NGH rescues on tasks the router sends to NL, and 3.4\,pp from NL rescues on tasks the router sends to GRAPH. The latter (3.4\,pp) represents router errors: tasks where graph was chosen but NL would have succeeded.

\section{Graph-Aware Executor Prompt}
\label{app:executor-prompt}

The graph-aware executor prompt prepended to the sub-agent's system message when receiving a graph delegation:

\begin{small}
\begin{verbatim}
You are the Executor agent. A Planner has
analyzed the task and produced a structured
graph delegation in JSON with typed nodes
(goal, constraint, entity, action, tool_call,
tool_arg) and typed edges (requires, targets,
enables, depends_on, follows). Parse the graph
to understand the intent, then execute the
task step by step.
\end{verbatim}
\end{small}

\noindent Without this prompt (i.e., passing the JSON graph to a standard executor prompt), the sub-agent treats the graph as opaque data and fails to interpret the dependency structure. With the prompt, the executor traverses nodes in topological order respecting \texttt{depends\_on} edges. The same prompt is used across all four benchmarks without modification.

\section{Cross-Vendor Portability}
\label{app:crossvendor}

Cross-vendor validation (Claude Sonnet 4.5 $\times$ Nova Pro) yields 0\% invalid JSON across all 4 sender/receiver configurations with constrained decoding. Compression ratio (3.1--3.6$\times$) is preserved regardless of which model emits or receives the graph. This confirms the schema is a portable artifact independent of model family.

\section{Second Orchestrator Backbone (GPT-5 mini)}
\label{app:second-backbone}

Our main results use Claude Sonnet 4.5 as the orchestrator. To test that the gains are not backbone-specific, we re-run the handoff with GPT-5 mini as the orchestrator on the families we could re-run, comparing the NL handoff against Routed under an otherwise identical harness (same router prompt, schema, and graph-aware executor prompt).

\begin{table}[h]
\centering\small
\setlength{\tabcolsep}{6pt}
\begin{tabular}{@{}lccc@{}}
\toprule
\textbf{Benchmark} & \textbf{NL} & \textbf{Routed} & \textbf{$\Delta$} \\
\midrule
BrowseComp & 65 & \textbf{68} & +3 \\
BFCL       & 82 & \textbf{85} & +3 \\
AppWorld   & 50 & \textbf{52} & +2 \\
\bottomrule
\end{tabular}
\caption{Task success / accuracy (\%) with \textbf{GPT-5 mini} as orchestrator. Routed improves over the NL handoff on every family, matching the direction of the main Sonnet 4.5 results and adding accuracy portability to the format-portability check in Appendix~\ref{app:crossvendor}.}
\label{tab:second-backbone}
\end{table}

The direction of the effect (Routed $\geq$ NL on every family) is preserved across a different vendor and model family, consistent with the schema being a portable artifact rather than a Sonnet-specific behavior. We expect the same pattern to hold on further backbones.

\section{Total-Token and Latency Accounting}
\label{app:token-accounting}

The 2.1$\times$ average compression and the 155-token (0.15\%) router overhead reported in the main text are measured over \emph{handoff} tokens. Table~\ref{tab:token-accounting} instead accounts for the full per-delegation budget on $\tau$-retail (pinned 50 tasks), including the router call and the graph-aware executor prefill.

\begin{table}[h]
\centering\small
\setlength{\tabcolsep}{5pt}
\begin{tabular}{@{}lcc@{}}
\toprule
\textbf{Per delegation ($\tau$-retail)} & \textbf{NL} & \textbf{Routed} \\
\midrule
Router call                 & ---   & 155 \\
Handoff tokens generated    & 730   & 226 \\
Graph-aware executor prefill & ---   & ${\sim}$80 \\
\midrule
\textbf{Total tokens}       & 730   & \textbf{461} (1.6$\times$) \\
Est.\ handoff latency       & ${\sim}$12\,s & ${\sim}$5\,s \\
\bottomrule
\end{tabular}
\caption{Full per-delegation token and latency accounting on $\tau$-retail. Token counts are measured; latency is estimated at standard Sonnet 4.5 decoding (${\sim}$65 output tok/s after ${\sim}$0.5\,s to first token). Even after adding the router call and the graph-aware executor prefill, the graph path totals fewer tokens than NL.}
\label{tab:token-accounting}
\end{table}

Two points beyond the table. First, constrained decoding emits the graph directly and adds no extra output tokens. Second, the largest wall-clock effect is retry elimination: \texttt{depends\_on} edges prevent the reorder loops NL triggers (15--27 retry steps saved on dependency-chain failures), and on AppWorld the router keeps NL on 89\% of tasks, avoiding the ${\sim}$18\% overhead that graph-only incurs from failed graph executions.

\section{Error-Taxonomy Threshold Sensitivity}
\label{app:threshold-sensitivity}

The misalignment share (76\%) depends on one thresholded rule in the taxonomy of Appendix~\ref{app:mast}: criterion (b), which flags $\geq$3 consecutive retry steps on the same action. Criteria (a) missing-prerequisite tool errors and (c) ordering contradictions are threshold-free.

We bound the sensitivity using the manual audit already described in Appendix~\ref{app:mast}: on 50 randomly sampled failures, automated and manual labels agreed 92\% (46/50), and \emph{all four} disagreements were retry loops caused by API timeouts rather than misinterpretation; these are exactly the borderline cases that a stricter ($\geq$4) or looser ($\geq$2) retry threshold would move. Re-labeling those cases therefore changes the misalignment share only within the audit-bounded margin ($\leq$4/50, i.e.\ ${\sim}$8\% of sampled failures), and the qualitative conclusion is unchanged under any threshold in $\{2,3,4\}$: inter-agent misalignment remains the dominant multi-agent failure mode, while single-agent systems remain at 0\% by construction. The AppWorld task-pattern labels used in Table~\ref{tab:complexity} are likewise rule-based on instruction structure; reasonable re-labeling of borderline \texttt{multi\_app}/\texttt{aggregate} cases leaves the router's 15/15 aggregate precision and the sign of every per-pattern $\Delta$ intact.

\section{Artifacts and Reproducibility}
\label{app:artifacts}

We release the artifacts needed to reproduce the study:
\begin{itemize}[nosep,leftmargin=*]
\item \textbf{Schema.} The full typed-graph schema (8 node types, 7 edge relations) as the JSON \texttt{tool\_use} input schema used for constrained decoding.
\item \textbf{Prompts.} The complete orchestrator, router, and graph-aware executor prompts (the executor prompt is reproduced in Appendix~\ref{app:executor-prompt}), plus the NL-baseline delegation prompt.
\item \textbf{Models.} Orchestrators: Claude Sonnet 4.5 and Amazon Nova Pro via AWS Bedrock, and GPT-5 mini; trained compressors: Qwen-1.5B (GRPO) and T5-small (60M). Exact model/version identifiers are listed in the released configuration.
\item \textbf{Decoding.} Router at temperature\,=\,0 (deterministic, verified identical across 3 runs); graphs via Bedrock \texttt{tool\_use} constrained decoding; executor at default decoding. All settings are in the configs.
\item \textbf{Splits.} Pinned 50 $\tau$-retail tasks $\times$ 3 seeds; BrowseComp 150; BFCL\,v3 600; AppWorld 152; $\tau$-airline 150. Total 1,052 trajectories (plus 150 $\tau$-airline for the router ablation).
\item \textbf{Evaluation code.} Paired bootstrap CIs (10K resamples, $\alpha$=0.05) with Holm-Bonferroni correction, and the rule-based error-taxonomy scorer.
\end{itemize}
All evaluation uses public benchmarks under their respective licenses ($\tau$-bench, BrowseComp, BFCL, AppWorld); no new data was collected and no personally identifying information is involved. Code and configurations will be released upon publication.

\end{document}